\documentclass{article}    

\usepackage[numbers]{natbib}
\usepackage[nonatbib,final]{nips_2018}

\usepackage[utf8]{inputenc} 
\usepackage[T1]{fontenc}    
\usepackage{hyperref}       
\usepackage{url}            
\usepackage{booktabs}       
\usepackage{amsfonts}       
\usepackage{nicefrac}       
\usepackage{microtype}      

\usepackage{amssymb}
\usepackage{enumitem}

\newcommand{\s}{\mathbf{s}}
\newcommand{\hats}{\mathbf{\hat s}}

\renewcommand{\description}[1]{}
\newcommand{\topic}[1]{}

\title{Adversarial Vision Challenge}

\author{
Wieland Brendel \textsuperscript{1},
Jonas Rauber \textsuperscript{1},
Alexey Kurakin \textsuperscript{2},
Nicolas Papernot \textsuperscript{3},
\\[0.2mm]
\textbf{
Behar Veliqi \textsuperscript{1},
Marcel Salath\'e \textsuperscript{4},
Sharada P. Mohanty \textsuperscript{4},
Matthias Bethge \textsuperscript{1}}
\\[0.5mm]
\textsuperscript{1}University of Tübingen,
\textsuperscript{2}Google Brain,
\textsuperscript{3}Pennsylvania State University,
\textsuperscript{4}EPFL
\\[0.5mm]
\texttt{adversarial-vision-challenge@bethgelab.org}
}

\begin{document}

\maketitle

\begin{abstract}
The NIPS 2018 Adversarial Vision Challenge is a competition to facilitate measurable progress towards robust machine vision models and more generally applicable adversarial attacks. This document is an updated version of our competition proposal that was accepted in the competition track of 32nd Conference on Neural Information Processing Systems (NIPS 2018).
\end{abstract}

\section{Overview of the competition}
\description{Summarize the background, available data, methods, available baseline, potential impact.}
This challenge is designed to facilitate measurable progress towards robust machine vision models and more generally applicable adversarial attacks.

Modern machine vision algorithms are extremely susceptible to small and almost imperceptible perturbations of their inputs (so-called \textit{adversarial examples}) \citep{biggio2013, szegedy13, goodfellow2014, papernot16}. This property reveals an astonishing difference in the information processing of humans and machines and raises security concerns for many deployed machine vision systems like autonomous cars. Improving the robustness of vision algorithms is thus important to close the gap between human and machine perception and to enable safety-critical applications.

In a robust network no attack should be able to find imperceptible adversarial perturbations. We thus propose to facilitate an open competition between neural networks and a large variety of strong attacks, including ones that did not exist at the time when the networks have been proposed. To this end the competition has one track for robust vision models as well as one track for targeted and one for untargeted adversarial attacks. Submitted models and attacks are continuously pitted against each other on an image classification task. Attacks are able to observe the decision of models on a restricted number of self-defined inputs in order to craft model-specific minimal adversarial examples.

\subsection{Keywords}
Robustness, adversarial attacks, adversarial examples, machine learning security

\subsection{Novelty}
\description{Have you heard about similar competitions in the past? If yes, describe the key differences. O/w disregard. Indicate whether this is a completely new competition, a competition part of a series, eventually re-using old data.}

We are aware of two related competitions, both of which were co-organised by authors of this proposal:

\begin{enumerate}
    \item \textbf{NIPS 2017 Competition on adversarial attacks and defenses.} \\
    \textit{Co-organised by Alexey Kurakin.}\\
    This competition pitted models against attacks but only indirectly: attacks were unable to query the models and hence had to device generic adversarial examples that would work against as many models as possible. Devising defenses against such unspecific transfer-based attacks is much simpler than becoming robust against model-specific attacks.
    \item \textbf{Robust Vision Benchmark (RVB).}\\
    \textit{Organised by Wieland Brendel, Jonas Rauber and Matthias Bethge.}\\
    The RVB is a continuously running benchmark (started in August 2017) in which submitted models are evaluated against a wide number of attacks (including submitted ones). Attacks are able to query the model both for confidence scores as well as gradients. This setting is interesting in order to evaluate model robustness but does not represent a realistic security scenario.
\end{enumerate}

This proposal can be seen as a follow-up to last year's NIPS competition but with a new concept to facilitate a direct co-evolution of robust vision models and more generally applicable adversarial attacks. By drawing lessons learnt from our previous competitions, we believe that this concept enables much better benchmarking of robustness and the results should be interesting both for computer vision and for the security of deployed machine learning systems.

\section{Competition description}

\subsection{Background and impact}
\description{Provide some background on the problem approached by the competition and fields of research involved. Describe the scope and indicate the anticipated impact of the competition prepared (economical, humanitarian, societal, etc.). Justify the relevance of the problem to the targeted NIPS community and indicate whether it is of interest to a large audience or limited to a small number of domain experts (estimate the number of participants). A good consequence for a competition is to learn something new by answering a scientific question or make a significant technical advance.}

\topic{Explaining adversarial examples}
One of the most striking differences between human and machine perception is the susceptibility of modern machine vision algorithms to extremely small and almost imperceptible perturbations of their inputs \citep{biggio2013, szegedy13, goodfellow2014, papernot16}. A tiny bit of carefully chosen image noise is usually sufficient to derail object detection with neural networks (e.g. flip the prediction from \emph{dog} to \emph{banana}). Such perturbations are commonly denoted as \emph{adversarial} and algorithms to find them are called \emph{adversarial attacks}. Adversarial perturbations reveal that decision making in current deep neural networks is based on correlational rather than causal features. From a security perspective they are worrisome because they open avenues to manipulate sensory signals in ways that go unnoticed for humans but seriously affect machine decisions.

\topic{Why are decision-based attacks important?}
So far, existing attacks (see \citep{brendel17, papernot17} for a taxonomy) had only limited success to threaten real-world applications like autonomous cars which do not convey internal model information like gradients or confidence values to an attacker. Even current transfer-based attacks can be effectively defended against through ensemble adversarial training, as was demonstrated in the NIPS 2017 competition. In addition, even if internal model information is available most existing attacks are easily disarmed through simple means like gradient masking or intrinsic noise. An important goal of this competition is to foster the development of stronger and more query-efficient attacks that do not rely on any internal model information but only on the final decision \citep{tramer17, brendel17}. These so-called \emph{decision-based} attacks have only recently been described for vision models \citep{tramer17, brendel17} but are highly relevant in real-world scenarios and much harder to defend than transfer-based or gradient-based attacks.

\topic{defenses/Robustness against adversarial examples}
Adversarial examples highlight that neural networks do not rely on the same causal features that humans use in visual perception. Closing this gap is important for many reasons: it would enable safety-critical applications of neural networks, would make neural networks more interpretable, would provide us with a deeper understanding of the human visual system and would increase the transferability of feature learning. Despite these advantages and many publications we are not aware of any significant progress towards more robust neural networks in complex visual tasks like object recognition, see also \citep{carlinigroundtruth, obfuscatedgradients}. A core problem is the proper evaluation of model robustness: a model might just be perceived to be robust because the attacks deployed against it fail. However, most attacks fail due to trivial side effects like gradient masking. Thus, just like in cryptography, the real test of model robustness is how well it stands against the scrutiny of attacks that are specifically designed against it. For this reason the competition is set up as a two-player game in which models and attacks are continuously pitted against each other. This encourages a co-evolution in which attacks can progressively adapt to the defense strategies of the models, and in which the models can progressively learn to better defend against strong attacks. Attacks will be able to query individual models, thereby enabling them to craft model-specific adversarials.

\topic{Desired outcome \& impact. Scientific questions / technical advance}
This competition seeks to stimulate progress towards more robust vision models as well as stronger and more generally applicable adversarial attacks. We believe that more robust models will require significant shifts in the type of representations that DNNs learn and the way they extract those features from their inputs. The results of this competition thus have the potential to impact many areas of machine learning. At the same time, stronger decision-based attacks will make it easier to evaluate the robustness of applied machine learning models and to assess under what circumstances these models fail.

\topic{Target audience / size}
In last year's NIPS competition 107 teams (187 people) participated in the defense track, 91 teams (153 people) in the non-targeted attack track and 65 teams (107 people) in the targeted attack track. Furthermore, the interest in adversarial robustness by the machine learning community has sharply risen throughout the last years: according to Google scholar the number of papers on \emph{adversarial examples} increased from 11 in 2014 to 64 (2015) to 167 (2016) to 612 (2017), which represents an 55-fold increase within four years. During the same period the number of publications on \emph{deep neural networks} increased only nine-fold (again according to Google scholar). In ICLR 2018 alone more than 40 papers were submitted that directly dealt with adversarial examples and defenses and another 55 papers mentioned them. It is thus safe to assume that this year's competition would gather even more participants than last year.

\subsection{Data}

\description{If the competition uses an evaluation based on the analysis of data,
please provide detailed information of the available data and
their annotations, and, in case, what the data generation
procedure will be (in this case, it must be clear in the document
that the data will be ready prior to the official launch of the
competition). Please justify that: (1) you have access to large
enough datasets to make the competition interesting and draw
conclusive results; (2) the data can be made freely available;(3) the ground truth has been kept confidential.}

While adversarial examples exist in many domains we here focus on vision or more specifically on the popular task of object recognition in natural images. To this end we rely on \textsc{Tiny ImageNet}. This data set is derived from the full ImageNet ILSVRC data set and has 100,000 images with size 64 x 64 pixels categorized into one of 200 classes. \textsc{Tiny ImageNet} is freely available and we will provide wrappers for TensorFlow and PyTorch to simplify access to the data. We expect models to be trained on \textsc{Tiny ImageNet} and we will provide several pre-trained baseline models. For testing and development we collect images ourself.

We split the collected images into 500 development images and 4,500 test images. The 500 development images will be released to participants to help during the development process. The test images are further split into 8 $\times$ 500 images for the intermediate evaluation of the methods and a hold-out set of 500 images for the final evaluation at the end of the competition. All images from the test set will be kept secret until after the end of the competition.

\subsection{Tasks and application scenarios}

\description{Describe the tasks of the competition and explain to which specific real-world scenario(s) they correspond to. If the competition does not lend itself
to real-world scenarios, provide a justification. Justify that the problem posed are scientifically or technically challenging but not impossible to
solve. If data are used, think of illustrating the same scientific problem using several datasets from various application domains.}

The competition will include three tasks:
\begin{enumerate}
    \item \textbf{Generate minimum untargeted adversarial examples.} In this task participants are given a sample image and access to a model. The goal is to create an adversarial image that is as similar as possible to the sample image (in terms of $L_2$ distance) but is wrongly classified by the given model.\\
    \textit{Application domain: This scenario corresponds to confusing a deployed machine learning model by preventing it from correctly identifying an object (e.g. not recognizing a STOP sign). In addition, the attack strategies will give us insights into how neural networks perceive the world and classify inputs.}
    \item \textbf{Generate minimum targeted adversarial examples.} In this task participants are given a sample image, a target class and access to a model. The goal is to create an adversarial image that is as similar as possible to the sample image (in terms of $L_2$ distance) but classified as the target class by the given model.\\
    \textit{Application domain: Same as above but with increased safety concerns (e.g. detecting a \$10 cheque as a \$1,000,000 cheque).}
    \item \textbf{Increase size of minimum adversarial examples.} In this task participants design robust object recognition models. For each given sample each adversarial attack proposes a minimum adversarial example for the given model. To goal of this task is to increase the size of the minimum adversarial perturbations (the $L_2$ distance between the best adversarial example and the sample image) the attacks can find.\\
    \textit{Application domain: any safety-critical vision application for which we need to ensure reliable visual inference.}
\end{enumerate}

Participants can submit solutions to all three tasks. All submissions are continuously pitted against each other on a fixed set of samples to encourage a co-evolution of robust models and better adversarial attacks.

\subsection{Metrics}

\description{For quantitative evaluations, select a scoring metric and justify
that it effectively assesses the efficacy of solving the problem
at hand. It should be possible to evaluate the results
objectively. If no metrics are used, explain how the evaluation
will be carried out.}

Participants can submit robust models as well as adversarial attacks. Models and attacks are pitted against each other on a number of image samples. Intuitively, we want the model score to represent the expected size of the minimum adversarial perturbations (larger is better). Conversely, the attack score should represent the expected size of the adversarial perturbations it can generate (smaller is better).

To be more precise we denote the set of submitted models as $M$, the set of attacks as $A$ (both targeted and untargeted) and the set of samples as $S$. The top-5 models and attacks are denoted as $M_5$ and $A_5$ respectively. We further denote an adversarial image $\hats$ for a given sample $\s\in S$ generated by attack $a\in A$ against model $m\in M$ as $\hats_a(\s, m)$. As a distance metric between to images $\s_1$ and $\s_2$ we use the $L_2$ distance,
\begin{equation}
    d(\s_1, \s_2) = \left\|\s_1 - \s_2\right\|_2.
\end{equation}

\paragraph{Model score.} For each model $m$ and each sample $\s$ we compute the adversarial $\hats_a(\s, m)$ for the attacks $a\in A_5$. We then determine the size of the smallest adversarial perturbation,
\begin{equation}
    d_m^{min}(\s, A_5) = \min_{a\in A_5} d(\s, \hats_{a}(\s, m)).
\end{equation}
If for a given sample $\s$ no attack was able to generate an adversarial example we set $d_m^{min}(\s)$ to a conservative upper bound. Finally, the model score is calculated as the median across the minimum distances,
\begin{equation}
    \mathrm{ModelScore}_m = \mathrm{median}(\left\{d_m^{min}(\s, A_5)|s\in S\right\}).
\end{equation}

\paragraph{Attack score.} We run each attack $a$ against the top-5 models $m\in M_5$ and each sample $\s\in S$. For each model and sample we compute the distance 
\begin{equation}
    d_a(\s, m) = d(\s, \hats_{a}(\s, m)).
\end{equation}
If the attack fails to generate an adversarial we set the corresponding distance to a conservative upper bound $d_a(\s, m)$. The final attack score is then the median size of the adversarial perturbation,
\begin{equation}
    \mathrm{AttackScore}_a = \mathrm{median}(\left\{d_a(\s, m)|\s\in S, m\in M_5\right\})
\end{equation}
The median is important to make the evaluation robust against outliers. For attacks lower scores are better, for models higher scores are better.

Both scores depend on the set of top-5 models and attacks. This focuses attacks on the hardest models and makes the evaluation feasible but also introduces a recursive dependence between (a) evaluating model/attack scores and (b) determining the top-5 in each track. This does not affect the final evaluation in which we pit all models against all attacks, which allows us to reliably determine the top-5 model and attack submissions (see \ref{sec:finaleval}). During the rest of the competition we determine the top-5 models \& attacks every two weeks (in the same way we perform the final evaluation) and all submissions will be tested against them until the next evaluation round.

\subsection{Baselines and code available}

\description{Specify what are (will be) the baselines for the competition, and
whether there is available code for participants (e.g., a starting
kit) and evaluation. Provide preliminary results, if available.}

\paragraph{Model baselines.} We will provide three baselines, all of which are based on the ResNet-50 model: (1) a vanilla model, (2) an adversarially trained model, and (3) a vanilla model with intrinsic frozen noise. We will provide the pretrained weights for all models.

\paragraph{Untargeted attack baselines.} We will provide five baselines: (1) a simple attack using additive Gaussian noise, (2) a simple attack using salt and pepper noise, (3) the Boundary Attack \citep{brendel17} with reduced number of iterations, (4) a single-step transfer attack and (5) an iterative transfer attack.

\paragraph{Targeted attack baselines.} We will provide four baselines: (1) a simple interpolation-based attack, (2) the Pointwise attack \citep{foolbox}, (3) the Boundary attack \citep{brendel17} and (4) an iterative transfer attack.

\subsection{Tutorial and documentation}

\description{Provide a reference to a white paper you wrote describing the
problem and/or explain what tutorial material you will provide.}

We will release an extensive development package containing test images, tutorials, example submissions and evaluation scripts:
\begin{itemize}
    \item Example model submission with placeholder for user-defined models (framework agnostic) and a tutorial on how to use it.
    \item Example attack submission with placeholder for user-defined adversarial attacks (framework agnostic) and a tutorial on how to use it.
    \item Code of all baseline attacks, including a detailed description.
    \item Code and model weights of all baseline models, including a detailed description.
    \item Set of 500 test images which participants can use for development of their models and attacks.
    \item Tool to evaluate model and attack submissions before the actual submissions. In this way users can test their code and its runtime behaviour before the actual submission.
    \item A reading list summarizing publications relevant for this competition.\footnote{\tiny{\url{https://medium.com/@wielandbr/reading-list-for-the-nips-2018-adversarial-vision-challenge-63cbac345b2f}}}
\end{itemize}
Some of the above mentioned code and tutorials will be reused and adapted from the NIPS 2017 competition, from the Robust Vision Benchmark as well as from previous competitions run by crowdAI.

\section{Organizational aspects}
\subsection{Protocol}

\description{Explain the procedure of the competition: what the participants will have to do, what will be submitted (results or code), and the evaluation procedure.
Will there be several phases? Will you use a competition platform with on-line submissions and a leader board? Indicate means of preventing cheating.
Provide your plan to organize beta tests of your protocol and/or platform.}

The competition will be hosted on crowdAI (https://crowdai.org). Participants submit their models and attacks as Docker images (see section \ref{sec:submission}). Submissions are continuously evaluated throughout the competition (see section \ref{sec:conteval}). The top-5 models and attacks against which submissions are tested will be determined every two weeks (see section \ref{sec:top5}), at which point all submissions are re-evaluated and the leaderboard is updated accordingly.

At the end of the competition we perform a final evaluation to determine the winners in each track (see section \ref{sec:finaleval}). The models and attacks are run in an isolated Kubernetes environment with local subnetworks that restrict intercommunication to the exchange between a single model and a single attack to prevent cheating. The communication is further restricted via a very limited HTTP API interface. To prevent models from memorising the correct labels and clean images (e.g. to perform a nearest-neighbour over the clean images), we test that models return decisions in a sequence-independent manner (i.e. model decisions should not depend on past inputs).

Before the official start of the competition we will simulate a full competition, including virtual participants and submissions, to test all parts of the system.

\subsubsection{Submission process}
\label{sec:submission}

Given the nature of the challenge participants are expected to package their models as docker images. Submissions are allowed in two ways: either as simple Docker image dumps or as code repositories.

\paragraph{Docker image dumps.} Participants build Docker images along given specifications (which we publish alongside some simple examples in the development package) and submit the final image to our servers. This option is most suitable for experienced users and guarantees maximum flexibility.

\paragraph{Code repositories.} To decrease the entrance barriers for participants not as comfortable with the Docker ecosystem, we allow simple code submissions based on Binder (https://mybinder.org/). Binder allows users to distill the software environment of their code as a set of configuration files in their source code from which we can deterministically generate a Docker image using \emph{repo2docker}. In the development package we will provide a series of template submission repositories which are already pre-configured with popular libraries of choice like Tensorflow. Participants can use the binder tools to locally test their code before making the submissions. The code repositories will be hosted on a custom gitlab instance on crowdAI.

By having submissions as repositories we will also ensure high reproducibility in the cumulative results of the challenge. At the end of the challenge the repositories will be made publicly accessible with the participant's open source license of choice (among those licenses referenced by the Open Source Initiative).

\subsubsection{Continuous evaluation}
\label{sec:conteval}

Participants can submit their models or attacks at any point in time. The number of submissions is limited to at most one submission per track within 24 hours. The submitted Docker images are evaluated in the backend against the top-5 opponents (either models or attacks depending on track) on 200 validation samples to determine the score for that submission.

\subsubsection{Top-5 evaluation round}
\label{sec:top5}

Every four weeks we perform a more extensive evaluations of all submissions (for each team the newest submission counts) to determine the new top-5 models and attacks. We use a test set of 200 secret sample images which are different in each evaluation round. The evaluation is performed according the following protocol:
\begin{itemize}
    \item The submission system is frozen for 48 hours during which the evaluation is performed.
    \item Round 1: All model/attack combinations are evaluated on a small set of 10 samples. From this evaluation we determine a very rough estimate of model and attack scores. Only the best 50\% are considered for the next round.
    \item Round 2: The remaining model/attack combinations are evaluated in the same way as in Round 1 but on a larger set of 20 samples. Again we determine the top 50\% of the remaining submissions.
    \item We iterate these rounds until we end up with the top-10 models and attacks. For these submissions we evaluate all model/attack combinations on the full test set of 200 samples to rigorously determine the top-5 submissions in each track.
    \item Scoring round: All submissions are re-scored on the 200 validation images and the leaderboards are updated accordingly.
\end{itemize}

\subsubsection{Final evaluation round}
\label{sec:finaleval}

To be scored in the final evaluation round participants have to release their code as open source. The scoring is performed in the same way as in the Top-5 evaluation rounds but this time the final scoring is performed on 500 secret test images. These test images have not been used in any of the evaluation rounds before.

\subsection{Rules}

\description{Provide a list of special rules.}

\begin{itemize}
    \item Bethgelab and Google Brain employees can participate but are ineligible for prices.
    \item To be eligible for the final scoring, participants are required to release the code of their submissions as open source.
    \item Any legitimate input that is not classified by a model (e.g. for which an error is produced) will be counted as an adversarial.
    \item If an attack fails to produce an adversarial (e.g. because it produces an error), then we will register a worst-case adversarial instead (a uniform grey image).
    \item Each classifier must be stateless and act one image at a time. This rule is supposed to prevent strategies such as memorizing pre-attack images and classifying replayed versions of them at defense time.
    \item The decision of each classifier must be deterministic. In other words, the classifier decision must be the same for the same input at any point in time.
    \item Attacks are allowed to query the model on self-defined inputs up to 1,000 times / sample. This limit is strictly enforced in the model/attack interface and an error will be returned whenever the attack queries the model more often.
    \item Each model has to process one image within 40ms on a K80 GPU (excluding initialization and setup which may take up to 100s).
    \item Each attack has to process a batch of 10 images within 900s on a K80 GPU (excluding initialization and setup which may take up to 100s).
\end{itemize}

\subsection{Schedule}

\description{Provide a time line for competition preparation and for running the competition itself. Propose a reasonable schedule leaving enough time for the organizers to prepare the event (a few months), enough time for the participants to develop their methods (e.g. 90 days), enough time for the organizers to review the entries, analyze and publish the results.}

Proposed schedule for the competition:
\begin{itemize}
    \item \textbf{April 20, 2018.} Launch website with announcement and competition rules. Start active advertisement of the competition.
    \item \textbf{June 18, 2018.} Release development kit for participants.
    \item \textbf{July 2 - November 1, 2018.} Competition is running. At the beginning, baselines will serve as top-5 models and attacks. Submissions are continously evaluated, the top-5 are determined every two weeks.
    \item \textbf{November 1, 2018} Deadline for the final submission.
    \item \textbf{November 1 - 15, 2018} Organizers evaluate submissions.
    \item \textbf{November 15, 2018} Announce competition results and release evaluation set of images.
\end{itemize}

\subsection{Competition promotion}

The competition will be promoted via the organisers' Facebook, Twitter, Google+ and Reddit accounts as well as the CleverHans blog, the crowdAI email list and several university mailing lists. Ian Goodfellow has also agreed to promote the challenge via his social media channels. Finally, we are currently in contact with Nvidia and Intel to secure prices for the top teams in each track.

\subsection{Organizing team}

\description{Provide a short biography of all team members, stressing their competence for their assignments in the competition organization. Make sure to include: coordinators, data providers, platform administrators, baseline method providers, beta testers, and evaluators.}

The proposed competition is technically challenging to run. It requires a complex orchestration to evaluate all submissions on tens of cloud GPU instances simultaneously throughout the whole competition. It requires extensive tooling to reduce the entrance barrier to participants as much as possible. And it requires much attention to the community and the platform to detect and resolve problems quickly. We have assembled a diverse team with recognised experts in the field of adversarial robustness to design, administrate and promote the competition as well as software and orchestration experts to design, implement and run the backend. Most members of the team have extensive experience with running large-scale machine learning competitions.

\vspace{0.5cm}

\noindent
\begin{minipage}[h][0.3cm][t]{0.6\textwidth}
\end{minipage}\hfill
\begin{minipage}[h][0.3cm][t]{0.38\textwidth}
\hspace{0.8cm}\emph{Roles}
\end{minipage}

\noindent
\begin{minipage}[h][4.8cm][t]{0.6\textwidth}
\paragraph{Wieland Brendel} is a senior postdoctoral researcher in the lab of Matthias Bethge at the University of T\"ubingen, Germany. He is co-organizer of the Robust Vision Benchmark, co-author of the Foolbox \citep{foolbox} (which provides framework-agnostic implementations of many adversarial attacks) and was the first to develop an effective decision-based attack at ImageNet scale (the Boundary attack \citep{brendel17}).
\end{minipage}\hfill
\begin{minipage}[h][4.8cm][t]{0.38\textwidth}
\begin{itemize}[topsep=0pt,itemsep=-1ex,partopsep=1ex,parsep=1ex]
    \item Lead coordinator
    \item Competition design
    \item Platform administrator
    \item Oversees evaluation
    \item Beta tester
    \item Advertisement
\end{itemize}
\end{minipage}

\noindent
\begin{minipage}[h][3.5cm][t]{0.6\textwidth}
\paragraph{Jonas Rauber} is a PhD student under the supervision of Matthias Bethge. Together with Wieland he co-organised the Robust Vision Benchmark, developed the Foolbox \citep{foolbox} as well as the Boundary Attack \citep{brendel17} and worked on the noise-robustness of deep neural networks \citep{geirhos17}.
\end{minipage}\hfill
\begin{minipage}[h][3.5cm][t]{0.38\textwidth}
\begin{itemize}[topsep=0pt,itemsep=-1ex,partopsep=1ex,parsep=1ex]
    \item Platform administrator
    \item Community support
    \item Baseline method provider
    \item Preparation of dev package
    \item Beta tester
\end{itemize}
\end{minipage}

\noindent
\begin{minipage}[h][3.4cm][t]{0.6\textwidth}
\paragraph{Alexey Kurakin} is a senior research software engineer at Google Brain. Alexey was the lead organiser in last year's NIPS competition. In his research work he demonstrated that adversarial examples can exist in the physical world \citep{kurakin16} and was the first to develop adversarial training at ImageNet scale \citep{kurakin16a}.
\end{minipage}\hfill
\begin{minipage}[h][3.4cm][t]{0.38\textwidth}
\begin{itemize}[topsep=0pt,itemsep=-1ex,partopsep=1ex,parsep=1ex]
    \item Data provider (test images)
    \item Baseline method provider
    \item Advise on challenge design
\end{itemize}
\end{minipage}

\noindent
\begin{minipage}[h][3.4cm][t]{0.6\textwidth}
\paragraph{Nicolas Papernot} is a PhD student in the lab of Patrick McDaniel and an intern at Google Brain. He is a developer and maintainer of the CleverHans security library \citep{cleverhans}, proposed the first practical decision-based attack for deep neural networks \citep{papernot17} and has published several other works on attacks and defenses.
\end{minipage}\hfill
\begin{minipage}[h][3.4cm][t]{0.38\textwidth}
\begin{itemize}[topsep=0pt,itemsep=-1ex,partopsep=1ex,parsep=1ex]
    \item Advertisement
    \item Advise on challenge design
\end{itemize}
\end{minipage}

\noindent
\begin{minipage}[h][4.7cm][t]{0.6\textwidth}
\paragraph{Marcel Salath\'e} is a digital epidemiologist working at the interface of population biology, computational sciences, and the social sciences. In the summer of 2015, Marcel became an Associate Professor at EPFL where he heads the Digital Epidemiology Lab. He is the co-founder of crowdAI (https://crowdai.org), a platform for organizing machine learning challenges and for encouraging reproducible research in Artificial Intelligence.
\end{minipage}\hfill
\begin{minipage}[h][4.7cm][t]{0.38\textwidth}
\begin{itemize}[topsep=0pt,itemsep=-1ex,partopsep=1ex,parsep=1ex]
    \item Advise on submission and evaluation backend
    \item Advertisement
    \item Advise on challenge design
\end{itemize}
\end{minipage}

\noindent
\begin{minipage}[h][5.1cm][t]{0.6\textwidth}
\paragraph{Sharada Prasanna Mohanty} is a PhD student in the laboratory of Prof. Marcel Salath\'e at \'Ecole Polytechnique F\'ed\'erale de Lausanne, Switzerland. His research is on exploiting numerous aspects of Machine Learning to help solve real world problems in Biology. He is co-founder of crowdAI (https://crowdai.org) and co-organized the NIPS 2017 Learning to Run Challenge. He will help design and implement the submissions and evaluation backend, 
and help shape the outline and rules of the competition.
\end{minipage}\hfill
\begin{minipage}[h][5.1cm][t]{0.38\textwidth}
\begin{itemize}[topsep=0pt,itemsep=-1ex,partopsep=1ex,parsep=1ex]
    \item Design \& implement submission and evaluation backend
    \item Advise on challenge design
\end{itemize}
\end{minipage}

\noindent
\begin{minipage}[h][4.2cm][t]{0.6\textwidth}
\paragraph{Behar Veliqi} is a senior research software engineer in the lab of Matthias. Behar holds a B.Sc. in Computer Science and has worked as a software engineer at United Internet and IBM for several years. He has extensive experience in scaling machine learning experiments on high-performance cluster environments and has extensively worked with databases and cluster management tools.
\end{minipage}\hfill
\begin{minipage}[h][4.2cm][t]{0.38\textwidth}
\begin{itemize}[topsep=0pt,itemsep=-1ex,partopsep=1ex,parsep=1ex]
    \item Platform administrator
    \item Community support
    \item Preparation of dev package
    \item Design \& implement submission and evaluation backend
    \item Evaluator
\end{itemize}
\end{minipage}

\noindent
\begin{minipage}[h][5.3cm][t]{0.6\textwidth}
\paragraph{Matthias Bethge} is professor at the University of Tübingen since 2009 and has served as area chair at several conferences including NIPS. Matthias has a strong track record on model comparison and benchmarking both w.r.t. unsupervised representation learning (e.g. \citep{Theis2015c, Sinz2009a}), and w.r.t. neural and behavioral models (e.g. \citep{Theis2016,Klindt2017a,Kuemmerer2015a,Kuemmerer2017a}). Since 2010 Matthias Bethge has been the director of the Bernstein center and since 2016 vice chair of the national Bernstein network. He has also served as general chair of the annual Bernstein meeting and is the head of the newly established Tuebingen AI Center (http://tue.ai). He is inventor of neural style transfer \citep{Gatys2016a} and co-founder of DeepArt UG (http://deepart.io) and Layer7 AI (http://layer7.ai).
\end{minipage}\hfill
\begin{minipage}[h][5.3cm][t]{0.38\textwidth}
\begin{itemize}[topsep=0pt,itemsep=-1ex,partopsep=1ex,parsep=1ex]
    \item Supervisor of strategic \\ concept
    \item Advise on all issues related to the challenge
\end{itemize}
\end{minipage}

\section{Resources}
\subsection{Existing resources, including prizes}

\description{Describe your resources (computers, support staff, equipment, sponsors, and available prizes).}

Amazon AWS will sponsor the necessary cloud computing resources to evaluate submissions (\$65,000 worth of cloud compute resources, see letter of support). Additional in-house GPU capacities as well as access to the SwissDataScienceCenter (https://datascience.ch/) are available to evaluate submissions.

\bibliographystyle{plain}
\bibliography{references}

\end{document}